# Towards Operationalizing Social Bonding in Human-Robot Dyads

Imran Khan[1]

[1] Division of Cognition and Communication, Department of Applied IT, University of Gothenburg, Sweden
imran.khan@ait.gu.se

**Abstract.** With momentum increasing in the use of social robots as long-term assistive and collaborative partners, humans developing social bonds with these artificial agents appears to be inevitable. In human-human dyads, social bonding plays a powerful role in regulating behaviours, emotions, and even health. If this is to extend to human-robot dyads, the phenomenology of such relationships (including their emergence and stability) must be better understood. In this paper, we discuss potential approaches towards operationalizing the phenomenon of social bonding between human-robot dyads. We will discuss a number of biobehavioural proxies of social bonding, moving away from existing approaches that use subjective, psychological measures, and instead grounding our approach in some of the evolutionary, neurobiological and physiological correlates of social bond formation in natural systems: (a) reductions in physiological stress (the "social buffering" phenomenon), (b) narrowing of spatial proximity between dyads, and (c) inter-dyad behavioural synchrony. We provide relevant evolutionary support for each proposed component, with suggestions and considerations for how they can be recorded in (real-time) human-robot interaction scenarios. With this, we aim to inspire more robust operationalisation of "social bonding" between human and artificial (robotic) agents.

## 1 Introduction

Human-robot interaction (HRI) literature is abundant with work and discussions related to notions of "bonding" or "attachments" with social robot partners. This is a natural and inevitable direction for the field to head in. The (implicit or explicit) assumption that these robots are to exist as *social* agents, rather than technological tools, that humans are to engage, interact and collaborate with over the long term, brings with it the requirement to untangle, amongst other things, the phenomenology of our social relationships: how they form, why they change over time, what the functionality, utility and consequences of such relationships are and, critically, how all of these components can be measured in a meaningful way.

To date, much of the work in HRI looking at notions of social attachment or bonding has borrowed perspectives, definitions, and measures that are (implicitly or explicitly) inspired by social psychology. Perhaps the most prominent of which is the Attachment Theory proposed by Bowlby and Ainsworth [1] describing the nature, functionality, and characteristics of social relationships with respect to infant-caregiver relationships. Four characteristics, coined *proximity maintenance, safe haven, secure base,* and *separation distress* [2], have been used as the foundation on which the formation of social "attachment" in human-robot dyads has been evaluated: oftentimes using post-interaction approaches [3]–[6], or questions related to trust, competency, or feelings of positive affect [7], [8] as proxies for "attachment".

Whilst Attachment Theory was initially inspired by ethology and has been instrumental in our understanding of how and why social relationships might form in human-human dyads, interpretations over time have sought to apply principles that were grounded primarily in *infant-caregiver* attachment theory to social relationships beyond childhood: not without critique [2]. Criticism has also been raised against its validity in human-robot relationships [9]. Though detailed discussion for this is beyond the scope of this short paper, we also propose that this problem has been conflated through inconsistent usage of terms, and the historical interchanging of an "attachment" (and subsequent attachment behaviours) and a "bond" [10].

HRI researchers should take note of such problems: here, we agree with [10] in that an *attachment* and a *bond* are qualitatively different phenomenaL a critical consideration in our studies of human-robot relationships.

Circling back to the ethological roots, we should note that we are not the only species to form social bonds beyond infancy. Instead, this phenomenon is seen across the animal kingdom, e.g. [11]–[13], corresponding with significant physiological health benefits associated with bonding [13]: evidence that social bonds provide a set of evolutionary fitness benefits that extends across the natural world. Furthermore, the (evolution of) neuropeptides such as oxytocin as vasopressin, have been heavily implicated in social bonding (and its prosocial behaviours and physiological benefits) across a range of human, and non-human animal species [11]. Thus, social bonding has a rich, evolutionary basis in humans and non-human animals, intertwined with neurobiological and physiological mechanisms, resulting in subtle, unconscious, and involuntary psychological, physiological, and behavioural effects that subjective reporting of experience alone may not be able to disentangle.

The challenge is thus: how can we measure the formation and strength of a social bond in human-robot dyads? If social robots are to, in part, complement or replace humans in their functionality as social agents, then we would expect that the functionality and consequences of bonding with humans would also be present in bonding with robots. Understanding this is of critical importance in contexts where social robots are to play an assistive role, e.g. in health-related applications, or in contexts where collaboration with them should be homologous to that with humans. Drawing from existing literature across various disciplines, we suggest that the evolutionary and neurobiological basis of social bonding can provide a set of involuntary and unconscious bonding proxies (or *signatures*) that may provide further insight into social bonding in human-robot dyads. We propose that this approach may provide an alternative method to operationalising this phenomenon, moving away from the strictly-psychological, subjective, post-hoc measures that are widely used.

Inspired by previous work done by [14] and [9], the goal of this paper is to discuss and inspire potential approaches towards operationalizing the phenomenon of social bonding---its formation, maintenance, and temporal changes---between human-robot dyads. We will discuss a number of biobehavioural proxies, grounded in evolutionary, neurobiological and physiological mechanisms associated with social bond formation in natural systems. To limit these measurements to those that can (theoretically) be quantified and evaluated in HRI scenarios, we focus on three key components of social bond formation: (a) reductions in physiological stress (the "social buffering" phenomenon), (b) narrowing of spatial proximity between dyads, and (c) inter-dyad behavioural synchrony. Where suitable, we will also integrate these three characteristics in with the previously-discussed attachment theory: we put forth that the two former components in the list above may have compatibility with the "safe haven" and "proximity-seeking" functions of Attachment Theory, respectively.

In this short paper, however, we will not propose a formalised model or equation that integrates all three measurements into a single value. Instead, we hope that it simply serves as a starting point (for HRI researchers) for discussing the use of these types of (evolutionary-inspired, biobehavioural or physiological) measures to approach this particular problem.

## 2    Biobehavioural Proxies of Social Bonding in Human-Robot Dyads

In the coming sections, we will focus on three physiological and behavioural aspects associated with social bond formation as outlined above. We will present a brief overview of evidence from ethology and neurobiology in support of these measures as quantifiable features of social bonding, and offer some suggestions for how they can be measured in the context of HRI studies. We propose that, in many instances, it is possible for these features to be measured during the course of (long-term) HRI studies. However, we also recognise that, due to practical constraints associated with in-the-lab vs. (quasi-)in-the-wild studies and other experimental constraints, our suggestions can, at best, be informative, but not prescriptive.

As we briefly discussed in the previous section, these three components also share in common a set of neuropeptides---oxytocin and vasopressin---which act on various neural circuitry that facilitates social bonding, stress reduction, approach and other prosocial behaviours [15], [16] and behavioural and physiological synchrony [15]. One may argue that these three components may simply be proxy measures of internal hormonal dynamics. Indeed, measurements of these hormones has been a widely-used method of evaluating positive social relationships in human and non-human animal literature [15]. We mention this simply to highlight a currently-underutilised approach for researchers interested in human interactions with social robots: the measuring of endogenous hormonal levels (such as oxytocin concentration) may present itself as a more reliable measure of social bonding and the physiological effects of social interaction with robot partners in the long term, and as an appropriate biomarker [17] of (perceived) social relationships.

## 2.1 Alleviating Physiological and Psychological Distress: The Social Buffering Phenomenon

It has been proposed that a core function of attachments to other social individuals is that they provide a *"safe haven"* during times of distress---physical or emotional stress, danger, or uncertainty coming from the environment---by alleviating or palliating the (perception) of negative physical, mental, or emotional stress [2]. However, the question must be asked: what, precisely, does a "safe haven" *do*? Zilcha-Mano et al. [18] have proposed that the physical or psychological presence of an attachment figure provides a "sense of removing distress and receiving comfort, encouragement, and support from the attachment figure." What remains to be understood is how a *mental sense* (of removing distress, comfort, etc) can be appropriately quantified in HRI contexts. Indeed, work from psychology that has attempted to do so has relied on subjective, post-hoc assessments via questionnaires related to, for example, perceived social support and well-being [19], and self-reported assessments of anxiety and avoidance [20]. Work in HRI has also relied heavily on subjective assessments (e.g. of anxiety reduction [21]) associated with the presence of the robot. We argue that these *a posteriori* reflections fail to adequately capture the real-time dynamics of stress-reduction: instead, adding an element of subjectivity, temporal delay and, thus, potential erroneous noise to evaluations of the effects of a robot's presence.

The notion of a "safe haven" may not be far removed from evolutionary theories of social support, and in particular a phenomenon known as "social buffering" [22], [23]. The social buffering theory proposes that the presence of a social partner, and in particular one with whom a social bond is shared, can dampen physiological responses to (physical and psychological) stressors---that is, (perceived) physical or psychological danger, uncertainty, or threat---in both the short and long-term, by moderating HPA activation (resulting in reduced secretion of hormones such as cortisol) as well as "cushioning" sympathetic nervous system activity. Such a phenomenon, it has been suggested [13], may be a critical evolutionary reason for the formation of social relationships and social support, owing to the maladaptive effects of chronic exposure (and responses to) stressors.

Evidence from this is strong in both the non-human literature---for instance, rats [24], prairie voles [25], primates [12], [26], guinea pigs [27], sheep and fishes [28]---as well as human children [29] and adults [16]. Furthermore, the effects of "social buffering" have been found to follow the strength of a bond between dyads: with more prominent, anxiolytic effects associated with how strong the relationship is considered to be [30]. In humans, this effect is not restricted to infant-caregiver bonds, but is flexible and dynamic across different life stages [31]. In that sense, the stress-reducing, *safe haven* function proposed by Attachment Theory may instead be thought of as a subset of the lifelong social buffering effects.

Therefore, grounding the notion of a "*sense of removing distress"* in a set of *physiological* measurements associated with the "social buffering" phenomenon can give us a starting point from which we can start quantifying the stress-reducing, anxiolytic effects associated with social bonding (i.e. with a robot). In addition, the relationship between these stress-reducing effects and the relative strength of a social bond in humans [30] allows these measurements of stress-reduction to not only inform us about the potential formation of a social bond, but also provide insight into the dynamic fluctuations of such a relationship. What we propose, then, is that observing the "social buffering" effect in human-robot dyads, particularly if it becomes more pronounced over time, can be an accurate approximation of the formation (and relative strength) of a social bond between human and robot actors.

## 2.2 Measuring "Social Buffering" in Human-Robot Interactions

If reductions in anxiety or distress is a key function of the presence of a social bond partner (as proposed by both the *safe haven* function of Attachment Theory [2] and the social buffering theory [22]) then what follows is that a reduction in physiological stress responses in the presence of another social individual, may be indicative of the formation (or strengthening) of a social bond. In line with the neurobiology literature [32], we can consider a stress response to be increased HPA axis activity, leading to increased secretion of cortisol, as well as heightened activity of the sympathetic nervous system (typically associated with the "fight or flight" response).

However, measuring changes in HPA activity, cortisol levels, and sympathetic nervous system arousal, at the very least, requires invasive measures of salivatory or urinary samples, or inserting electrodes to peripheral nerves [33]: an impractical notion for HRI researchers. Instead, real-time changes in various biosignals and behaviours can serve as appropriate approximators for HPA (via cortisol levels) and ANS activity in individuals.

Changes in cortisol activity have often been associated with changes in heart rate [34] and heart rate variability [35] (though this has been contested given temporal sensitivity of heart rate variability): both of which can be measured through non-invasive photoplethysmography (PPG) devices. Pupillary information, such as increased pupil size [36] and increases in scanning entropy or reductions in fixation periods [37], have also been associated with higher cortisol levels. Finally, electroencephalogram (EEG) devices (which are more feasible in HRI contexts in recent years given the development of minimally-invasive, more accessible EEG devices (e.g. [38])) can be used to measure localised brain activity, with frontal asymmetry being seen as the main candidate for a neural correlate of stress [39]. Other studies go further and suggest that frontal (right-to-left) asymmetries of specific power bands are more indicative of increased cortisol activity [40]. Autonomic nervous system activity, on the other hand, can be assessed through galvanic skin response (GSR) or electrodermal activity (EDA): that is, sweat gland activity measured both through skin conductance levels (SCL, measuring tonic activity) as well as skin conductance response (SCR, measuring phasic activity). SCR in particular is found to increase exclusively through arousal of the sympathetic, but not parasympathetic, nervous system. An additional consideration is the measurement of blood pressure (BP), with increases in BP associated with sympathetic nervous system activation. Thus, combining EDA, HRV, and BP may help "triangulate" ANS activity. For further information, we point readers to the detailed review by [41] on various stress-related biosignals.

Some of these measurements are not new to HRI researchers, having been previously used to measure "arousal" as part of the commonly-used circumplex model of affect [42]. What we propose here is that, according to the social buffering theory, *temporal changes in physiological stress,* which can be linked to the presence of a social other and which becomes more prominent over time, may be a strong indicator of a social bond being formed between the human and the robot. Despite potential criticisms regarding inter-individual differences between baselines of stress, or the difficulty in establishing good ground truths for "stress" [41] we argue that these are not concerns when treating stress as a continuous, rather than discrete, dimension over the long term. Comparisons are therefore made against previous experiences and interactions over time, rather than against a fixed baseline "ground truth".

## 2.3   Spatial Proximity between Social Actors

A second evolutionary feature of social bonding is related to the spatial proximity of individuals, i.e. actual physical distance between individuals. Starting from parent-infant attachments, this is intuitive: infants have limited mobility, and their survival necessarily requires mothers, as a physical resource, to be in close proximity to them [43], facilitating life-critical social interactions (e.g. feeding), and reducing distress [22]. This corresponds to the "proximity maintenance" function of social attachment in Attachment Theory [2]. Yet, narrow spatial proximity between individuals persists beyond infanthood, with emotional closeness and intimacy, rather than needing life-critical resources, being a modulator of distance between actors [44]. From an evolutionary standpoint, narrower spatial proximity throughout life affords improved salience of social signals from others, and opportunities to engage in prosocial activities (e.g. grooming/mating, resource-sharing, and thermoregulation).

Evidence from both animal and human literature supports this. Spatial proximity has been related to existing affiliations in titi monkey dyads [45], the strength of affiliation in squirrel monkeys [46], social bond strength in goats [47], horses [48] and chimpanzees [49]. In humans, dynamic changes in interpersonal distance (IPD) have been associated with changes in relationship between social individuals [50] and represents the nature and quality of social relationships [51]: the closer a relationship between social individuals, the narrower the spatial proximity between them. Studies in HRI have also found IPD between human and robot to be associated with experience and historical interactions [52] as well as affective attitudes [53] towards robots. Underpinning these behavioural effects, once again, are neural (e.g. limbic and subcortical structures related to affective and social space regulation [54]) and hormonal mechanisms (e.g. release of oxytocin, facilitating prosocial and anxiolytic effects [16]).

Particularly for humans, the Social Baseline Theory [55] proposes that social proximity poses evolutionary advantages related to metabolic resource management: social resources (i.e. other social agents) are bioenergetic resources which, through notions of load sharing, reduces the (perceived and actual) metabolic cost or effort associated with difficult tasks [56]. SBT suggests that this relates to the notion of an *expanded* self with familiar others (i.e. those with whom a pre-existing relationship exists), but not strangers.

In sum, our neural circuitry acts as though social partners with whom we are socially-bonded are part of our embodied sense of self.

While contextual factors related to sex, height, gender, social anxiety, and attachment styles mean that absolute distances between social actors cannot, in themselves, be explicitly definitive of social bonding, we propose that this type of measurement may, instead, be complementary to other more defining components of social bond formation (such as "social buffering") and can shed further insight into temporal changes of a social bond over time.

## 2.4  Measuring Social Proximity in HRI

For simplicity, we assume two things: (1) that social proximity is to be evaluated between two physically (rather than virtually) embodied agents (human and robot), and (2) that spatial proximity in HRI contexts is adjusted only by the human actor.

Several existing approaches to measuring spatial proximity are already used in the field of proxemics. In human-human interactions, a common method for categorising interpersonal distance are the four spatial zones proposed by Hall [57]. One challenge, however, is understanding how zones relating to human-human interaction can be appropriately applied to human-robot contexts, where the granularity of proximity would typically be much smaller. For instance, mean distances between human and robot was 35.1cm to 50cm in [53] whereas typical distances between humans extend beyond this (e.g. 45cm as the lower bound on "personal space", and 3.6m as the upper limit on "social space" [57]). Other approaches have used the Euclidean distance (or similar) between social agents [58]: simply, the straight-line distance between two agents. Here, the most ecologically-valid approach is for researchers to allow the human-robot interaction scenario to play out in a natural manner, recording the interaction (via the use of (high-resolution) cameras, e.g. [59]) or via motion capture or recording of skeletal positions (e.g. through a Microsoft Kinect, as used in [60] along with the necessary computer vision models). Calculating the (average or absolute) distance that the human actor maintains over the course of the interaction can then be done in real-time (e.g. via computer vision models [61]) or in a post-hoc analysis (e.g. [62]).

Other work in HRI has used the "stop distance" or "approach distances" approaches (also known as "robot-to-human" or "human-to-robot" approaches, respectively): with the moving agent being either the robot or the human, and the human declaring when the mover was at a comfortable distance (e.g. [63]). However, these methods are suggested to be inconsistent and unreliable [64]. We suggest that this might be due to the lack of ecological validity: with both methods, participants necessarily need to be actively conscious of spatial proximity, rather than it being an unconscious and involuntary component of natural social interaction. We therefore suggest the more (ecologically-valid) approach set out above.

There are some additional considerations. The granularity of dynamic changes in spatial proximity in HRI scenarios is much smaller than in naturalistic, human-human or animal-animal interactions, requiring a set of hardware or software that is both accurate and sensitive to small changes. There are further considerations of what body parts distance should be measured between. Other inter-individual differences (e.g. sex, height, levels of social anxiety, and attachment styles [50]), as well as cultural and social norms, may also be confounding factors in assessments of spatial proximity. However, following from the animal literature, we emphasise again that the focus should be on inter-individual, inter-session differences---the *narrowing of spatial proximity over multiple sessions by the same human actor,* rather than a focus on absolute, arbitrary measurements, that can help determine the strength of a social bond between human and robot.

## 2.5  Inter-Dyad (Behavioural) Synchrony

The final component we will focus on is the social phenomenon of behavioural synchrony. Behavioural synchrony can be thought of as a "tendency for behavioural patterns to become more similar while two or more [people] interact" [65], or "...as a match between the interaction partners' behaviours in time" [66]. Seen as a phenomenon of an *extended* self with social partners [67] that starts in infancy [68], synchronisation of behaviour (as well as physiology) is also associated with increased levels of trust [69], empathy [70], cooperation [71], shared affect [72] and, crucial to our position, increased levels of affiliation and social bonding [65], [73]. As with the previous components of social bonding above, behavioural synchrony is associated with increases in circulating oxytocin, likely by acting on predictive neural circuitry [74] (e.g. of the social environment) or, as a proxy for grooming behaviours, acting on endogenous opioid systems [75]. Thus, behavioural synchrony, through the release of numerous endorphins and neurohormones that influence (and are influenced) by the formation of social bonds [75], intertwined with metabolic advantages and increases in cooperation and coordination between dyads, has played an important evolutionary role in the formation and maintenance of social connections.

These types of subtle and involuntary synchrony of bodily (micro-)behaviours and/or actions have been found to occur in several ways: such as synchronised eye gaze or movement, head movement, but also affective expression, body orientation, and the quality of vocal interaction between (e.g. matching of vocal pitch) [68]. The prevalence of dyadic synchrony has been found to be associated with the quality of the attachment in infant-caregiver [76] and adult-adult human dyads, or related to an *intention* to form a bond [77] with a social other. It has been suggested, therefore, that the quality of a relationship is embodied through the amount of timely coordinated movement between dyads: in many cases, the more synchrony, the stronger the social bond between them.

Though human-robot behavioural synchrony is not widely studied, some work coming from human-animal literature may provide evidence of the potential for cross-"species" behavioural synchrony. For instance, longitudinal evidence in human children (with autism spectrum disorder) interacting with dogs shows movement synchrony to increase in humans with respect to the dogs' movements over repeated interactions over time [78]. Similarly, eye gaze, joint attention, and object touch synchrony was found to emerge between dogs and human handlers in [79]. These studies, despite being in natural dyads, may form the foundations for understanding this phenomenon in natural-artificial system dyads.

## 2.6  Measuring Behavioural Synchrony in Human-Robot Dyads

Unlike the measures of stress reduction and physical proximity between human and robot described in previous sections, the notion of synchrony in HRI scenarios is, arguably, more difficult to evaluate and quantify. There are (at least) two reasons for this. Firstly, robots, even those with humanoid embodiments, do not possess the same freedom of movement, morphologies, or complete set of (micro)behaviours as humans. How do we assess synchrony between actors whose embodiments and afforded set of (micro)behaviours are fundamentally different? Secondly, approaches to measuring synchrony in human-human interaction have often relied on effortful ethnographic approaches: manual identification and coding of synchronised movement or activity between dyads by experimenters conducting the studies.

We suggest that both of these challenges can, theoretically, be overcome. Firstly, it is not a requirement for the entire repertoire of behaviours to be synchronised for "behavioural synchrony" to be sufficiently recorded. Instead, limiting observations of synchrony to modalities which can be reasonably mapped between human and robot agents might be sufficient. Initially, this may be limited to robots that have some humanoid-like embodiments, i.e. those that possess human-like (micro)behavioural qualities. As an example, the humanoid robot head Furhat [80] possess capabilities for head movement (including nodding, moving to face something, and head tilting), and face gestures (e.g. smiling or frowning) or micro-behaviours such as naturalistic eye-blinking: robots such as Nao [81] possess higher degrees of skeletal movement (including limbs and body), but no ability to express face gestures. Both robots possess voice capabilities. Thus, in this example, eye blink rates, body gestures, and vocal pitch of human participants can be mapped onto the corresponding behaviours of the robots, respectively, as a micro-behaviour that can be synchronised between both agents.

While several approaches to measuring behavioural synchrony have been used in the human and animal literature, we suggest that the most feasible approach in HRI contexts is via ethnographic methods. Here, researchers may leverage a combination of hardware and software to record behaviours (e.g. webcams to detect face or body movements, eye trackers for eyes; high-resolution camera for skeletal movement) or log digital events (i.e. eye blinks or kinematic data coming from the robot), review recordings/data logs post-study and manually encode behaviours (such as the behaviour(s) of interest, and time stamps) and thus calculating synchrony between parties. Despite algorithmic approaches to synchrony detection [82], the "gestalt-like" quality of behaviour synchrony means that ethnographic approaches are still the most successful and accurate method to use, oftentimes utilising multiple coders. One advantage in HRI contexts is the implied unidirectionality of synchrony: it is only the human's behaviour which is of interest (with the robot being the "leader" or "generator", and the human is the "follower" or "receiver"), thus reducing the overall effort required to calculate synchrony between partners.

To this end, several metrics for behavioural synchrony calculation have been proposed. Given the breadth of potential calculations, we will instead point readers to extended discussions found at [83], [84]. We will note, however, several factors that should be considered in HRI contexts. Firstly, since behavioural synchrony is a non-linear time-locked phenomenon, and that there will be an inherent lag between robot and human displays of behaviour, the first hurdle is to identify a suitable time window between behaviours for which to attribute behavioural synchrony to: one starting point might be the 5-10s window proposed by [85]. Then, the modality of behaviour being observed (e.g. auditory, motor/bodily, visual) brings different lower and upper bounds on behaviour synchrony [83]. Finally, given the difference in embodiments between humans and robots, researchers may benefit from identifying and exploring variations or other non-obvious behaviours that may correspond to synchrony in behaviours between agents. For example, humans and dogs do not locomote or raise their limbs in the same manner, yet behavioural synchrony in [78] is still reported at the level of lifting a leg or moving in a given direction. This study may inspire new ways to assess behavioural synchrony in human-robot dyads.

We recognise, however, that identifying and measuring synchrony between human and artificial agents poses a significant challenge in comparison to our previously-proposed methods. Whatever tools, approaches and metrics are used, we suggest, once again, that emphasis should be placed on the changes in these measures over time (i.e. with repeated exposure to the same robot agent) as a means to assess changes in the human's (perception of) bonding with the social robot.

## 3      Concluding Remarks

In this paper, we have presented three biobehavioral proxies of social bonds that we propose to the HRI community as potential viable measures of social bond formation in human-robot dyads. We argue that social bonds, with their rich, evolutionary history and functionality in both human and non-human animals, can provide crucial detectable and quantifiable signatures for the formation of social bonds between humans and autonomous social agents (i.e. robots).

However, we do not prescribe explicit, definitive ways in which can or should be measured. As we touch on throughout the paper, several measurement approaches may be valid, and this will be researchers' available tools and expertise. Nor do we currently propose a formal model for how to integrate all of these measures. Given the evolutionary relevance of these three components of social support, however, we suggest that experimenters should aim to record measurements across as many of these dimensions as possible. Else, we propose prioritising measurements in the order that we have presented them: changes in physiological stress, then (changes in) spatial proximity between human and robot. Due to its effortful nature---with the only feasible approach we identified currently being through ethnographic-inspired, coder ratings---we propose that identifying inter-dyad synchrony to be the lowest-priority component.

We note again that this list is purposefully open for interpretation and adjustment, and is also non-exhaustive. We have limited our present discussion in this short article to data points that we believe can be feasibly captured by HRI researchers in laboratory (and even quasi-in-the-wild) contexts. The purpose of this article was not to provide an exhaustive list of all biobehavioural correlates of social bond formation, but to open up the conversation of focusing on evolutionary (mechanisms) of social bond formation. Though this is not prescriptive, we suggest that researchers interested in notions of social bond formation or "attachment" in human-robot dyads can look to these types of measurements when operationalizing social bonds. Despite some of the relative challenges and caveats that we discuss, we believe that this type of approach will be crucial to our understanding of bonding and attachments between human and artificial social agents. In the near future, we aim to validate the feasibility and accuracy of these proposed approaches in several longitudinal HRI observations, and to extend this discussion to also propose a formal operationalisation for measuring the formation and quality of social bonds in human-robot dyads.

**References**

[1]   J. Bowlby, 'The Bowlby-Ainsworth attachment theory', *Behav. Brain Sci.*, vol. 2, no. 4, pp. 637–638, Dec. 1979, doi: 10.1017/S0140525X00064955.

[2]   C. Hazan and D. Zeifman, 'Pair bonds as attachments', *Handb. Attach. Theory Res. Clin. Appl.*, pp. 336–354, 1999.

[3]   R. Andreasson, B. Alenljung, E. Billing, and R. Lowe, 'Affective Touch in Human–Robot Interaction: Conveying Emotion to the Nao Robot', *Int. J. Soc. Robot.*, vol. 10, no. 4, pp. 473–491, Sep. 2018, doi: 10.1007/s12369-017-0446-3.

[4]   C. J. A. M. Willemse and J. B. F. van Erp, 'Social Touch in Human–Robot Interaction: Robot-Initiated Touches can Induce Positive Responses without Extensive Prior Bonding', *Int. J. Soc. Robot.*, vol. 11, no. 2, pp. 285–304, Apr. 2019, doi: 10.1007/s12369-018-0500-9.

[5]   F. Krueger, K. C. Mitchell, G. Deshpande, and J. S. Katz, 'Human–dog relationships as a working

framework for exploring human–robot attachment: a multidisciplinary review', *Anim. Cogn.*, vol. 24, pp. 371–385, 2021.

[6] T. Xie and I. Pentina, 'Attachment theory as a framework to understand relationships with social chatbots: a case study of Replika', 2022.

[7] O. Gillath, T. Ai, M. S. Branicky, S. Keshmiri, R. B. Davison, and R. Spaulding, 'Attachment and trust in artificial intelligence', *Comput. Hum. Behav.*, vol. 115, p. 106607, Feb. 2021, doi: 10.1016/j.chb.2020.106607.

[8] C. Di Dio et al., 'Shall I Trust You? From Child–Robot Interaction to Trusting Relationships', *Front. Psychol.*, vol. 11, p. 469, Apr. 2020, doi: 10.3389/fpsyg.2020.00469.

[9] E. C. Collins, A. Millings, and T. J. Prescott, 'Attachment to assistive technology: a new conceptualisation', in *Assistive technology: From research to practice*, IOS Press, 2013, pp. 823–828.

[10] K. W. Watson, 'Bonding and Attachment in Adoption: Towards Better Understanding and Useful Definitions', *Marriage Fam. Rev.*, vol. 25, no. 3–4, pp. 159–173, Sep. 1997, doi: 10.1300/J002v25n03_03.

[11] O. J. Bosch and L. J. Young, 'Oxytocin and Social Relationships: From Attachment to Bond Disruption', *Curr. Top. Behav. Neurosci.*, vol. 35, pp. 97–117, 2018, doi: 10.1007/7854_2017_10.

[12] C. Crockford, R. M. Wittig, K. Langergraber, T. E. Ziegler, K. Zuberbühler, and T. Deschner, 'Urinary oxytocin and social bonding in related and unrelated wild chimpanzees', *Proc. R. Soc. B Biol. Sci.*, vol. 280, no. 1755, p. 20122765, Mar. 2013, doi: 10.1098/rspb.2012.2765.

[13] J. Holt-Lunstad, T. B. Smith, and J. B. Layton, 'Social Relationships and Mortality Risk: A Meta-analytic Review', *PLoS Med.*, vol. 7, no. 7, p. e1000316, Jul. 2010, doi: 10.1371/journal.pmed.1000316.

[14] N. Rabb, T. Law, M. Chita-Tegmark, and M. Scheutz, 'An Attachment Framework for Human-Robot Interaction', *Int. J. Soc. Robot.*, vol. 14, no. 2, pp. 539–559, Mar. 2022, doi: 10.1007/s12369-021-00802-9.

[15] R. Feldman, 'Oxytocin and social affiliation in humans', *Horm. Behav.*, vol. 61, no. 3, pp. 380–391, Mar. 2012, doi: 10.1016/j.yhbeh.2012.01.008.

[16] M. Heinrichs, T. Baumgartner, C. Kirschbaum, and U. Ehlert, 'Social support and oxytocin interact to suppress cortisol and subjective responses to psychosocial stress', *Biol. Psychiatry*, vol. 54, no. 12, pp. 1389–1398, Dec. 2003, doi: 10.1016/S0006-3223(03)00465-7.

[17] I. Khan and L. Cañamero, 'The Long-Term Efficacy of "Social Buffering" in Artificial Social Agents: Contextual Affective Perception Matters', *Front. Robot. AI*, vol. 9, p. 699573, Sep. 2022, doi: 10.3389/frobt.2022.699573.

[18] S. Zilcha-Mano, M. Mikulincer, and P. R. Shaver, 'An attachment perspective on human–pet relationships: Conceptualization and assessment of pet attachment orientations', *J. Res. Personal.*, vol. 45, no. 4, pp. 345–357, Aug. 2011, doi: 10.1016/j.jrp.2011.04.001.

[19] S. C. Langston, *Understanding and quantifying the roles of perceived social support, pet attachment, and adult attachment in adult pet owners' sense of well-being*. Washington State University, 2014.

[20] N. L. Collins and B. C. Feeney, 'A safe haven: an attachment theory perspective on support seeking and caregiving in intimate relationships.', *J. Pers. Soc. Psychol.*, vol. 78, no. 6, p. 1053, 2000.

[21] C. S. Song and Y.-K. Kim, 'The role of the human-robot interaction in consumers' acceptance of humanoid retail service robots', *J. Bus. Res.*, vol. 146, pp. 489–503, Jul. 2022, doi: 10.1016/j.jbusres.2022.03.087.

[22] T. Kikusui, J. T. Winslow, and Y. Mori, 'Social buffering: relief from stress and anxiety', *Philos. Trans. R. Soc. B Biol. Sci.*, vol. 361, no. 1476, pp. 2215–2228, Dec. 2006, doi: 10.1098/rstb.2006.1941.

[23] M. R. Gunnar, 'Social Buffering of Stress in Development: A Career Perspective', *Perspect. Psychol. Sci.*, vol. 12, no. 3, pp. 355–373, May 2017, doi: 10.1177/1745691616680612.

[24] D. Suchecki, P. Rosenfeld, and S. Levine, 'Maternal regulation of the hypothalamic-pituitary-adrenal axis in the infant rat: the roles of feeding and stroking', *Dev. Brain Res.*, vol. 75, no. 2, pp. 185–192, Oct. 1993, doi: 10.1016/0165-3806(93)90022-3.

[25] M. Donovan, Y. Liu, and Z. Wang, 'Anxiety-like behavior and neuropeptide receptor expression in male and female prairie voles: The effects of stress and social buffering', *Behav. Brain Res.*, vol. 342, pp. 70–78, Apr. 2018, doi: 10.1016/j.bbr.2018.01.015.

[26] R. M. Wittig, C. Crockford, A. Weltring, K. E. Langergraber, T. Deschner, and K. Zuberbühler, 'Social support reduces stress hormone levels in wild chimpanzees across stressful events and everyday affiliations', *Nat. Commun.*, vol. 7, no. 1, p. 13361, Nov. 2016, doi: 10.1038/ncomms13361.

[27] M. B. Hennessy, R. Zate, and D. S. Maken, 'Social buffering of the cortisol response of adult female

guinea pigs', *Physiol. Behav.*, vol. 93, no. 4–5, pp. 883–888, Mar. 2008, doi: 10.1016/j.physbeh.2007.12.005.

[28] K. M. Gilmour and B. Bard, 'Social buffering of the stress response: insights from fishes', *Biol. Lett.*, vol. 18, no. 10, p. 20220332, Oct. 2022, doi: 10.1098/rsbl.2022.0332.

[29] M. Nachmias, M. Gunnar, S. Mangelsdorf, R. H. Parritz, and K. Buss, 'Behavioral Inhibition and Stress Reactivity: The Moderating Role of Attachment Security', *Child Dev.*, vol. 67, no. 2, p. 508, Apr. 1996, doi: 10.2307/1131829.

[30] C. D. Calhoun, S. W. Helms, N. Heilbron, K. D. Rudolph, P. D. Hastings, and M. J. Prinstein, 'Relational victimization, friendship, and adolescents' hypothalamic–pituitary–adrenal axis responses to an in vivo social stressor', *Dev. Psychopathol.*, vol. 26, no. 3, pp. 605–618, Aug. 2014, doi: 10.1017/S0954579414000261.

[31] J. R. Doom, C. M. Doyle, and M. R. Gunnar, 'Social stress buffering by friends in childhood and adolescence: Effects on HPA and oxytocin activity', *Soc. Neurosci.*, vol. 12, no. 1, pp. 8–21, Jan. 2017, doi: 10.1080/17470919.2016.1149095.

[32] B. Chu, K. Marwaha, T. Sanvictores, and D. Ayers, 'Physiology, Stress Reaction', in *StatPearls*, Treasure Island (FL): StatPearls Publishing, 2023. Accessed: Aug. 11, 2023. [Online]. Available: http://www.ncbi.nlm.nih.gov/books/NBK541120/

[33] E. Eatough, K. Shockley, and P. Yu, 'A review of ambulatory health data collection methods for employee experience sampling research', *Appl. Psychol.*, vol. 65, no. 2, pp. 322–354, 2016.

[34] G. Giannakakis et al., 'Stress and anxiety detection using facial cues from videos', *Biomed. Signal Process. Control*, vol. 31, pp. 89–101, 2017.

[35] T. F. of the E. S. of C. the N. A. S. of P. Electrophysiology, 'Heart rate variability: standards of measurement, physiological interpretation, and clinical use', *Circulation*, vol. 93, no. 5, pp. 1043–1065, 1996.

[36] P. Ren, A. Barreto, Y. Gao, and M. Adjouadi, 'Affective assessment by digital processing of the pupil diameter', *IEEE Trans. Affect. Comput.*, vol. 4, no. 1, pp. 2–14, 2012.

[37] L. Fridman et al., 'What can be predicted from six seconds of driver glances?', in *Proceedings of the 2017 CHI Conference on Human Factors in Computing Systems*, 2017, pp. 2805–2813.

[38] 'Muse: EEG-Powered Meditation & Sleep Headband'. https://choosemuse.com/ (accessed Aug. 21, 2023).

[39] A.-M. Brouwer, M. A. Neerinex, V. Kallen, L. van der Leer, and M. ten Brinke, 'EEG alpha asymmetry, heart rate variability and cortisol in response to virtual reality induced stress.', *Cyberpsychology Behav. Soc. Netw.*, vol. 4, no. 1, pp. 27–40, 2011.

[40] G. Giannakakis, D. Grigoriadis, and M. Tsiknakis, 'Detection of stress/anxiety state from EEG features during video watching', in *2015 37th Annual International Conference of the IEEE Engineering in Medicine and Biology Society (EMBC)*, IEEE, 2015, pp. 6034–6037.

[41] G. Giannakakis, D. Grigoriadis, K. Giannakaki, O. Simantiraki, A. Roniotis, and M. Tsiknakis, 'Review on Psychological Stress Detection Using Biosignals', *IEEE Trans. Affect. Comput.*, vol. 13, no. 1, pp. 440–460, Jan. 2022, doi: 10.1109/TAFFC.2019.2927337.

[42] Jonathan Posner, James A Russell, and Bradley S Peterson, 'The circumplex model of affect: an integrative approach to affective neuroscience, cognitive development, and psychopathology', *Dev. Psychopathol.*, 2005, doi: 10.1017/s0954579405050340.

[43] S. W. Porges, 'Social Engagement and Attachment', *Ann. N. Y. Acad. Sci.*, vol. 1008, no. 1, pp. 31–47, Dec. 2003, doi: 10.1196/annals.1301.004.

[44] N. L. Collins and S. J. Read, 'Adult attachment, working models, and relationship quality in dating couples.', *J. Pers. Soc. Psychol.*, vol. 58, no. 4, pp. 644–663, 1990, doi: 10.1037/0022-3514.58.4.644.

[45] E. Fernandez-Duque, C. R. Valeggia, and W. A. Mason, 'Effects of Pair-Bond and Social Context on Male-Female Interactions in Captive Titi Monkeys (Callicebus moloch, Primates: Cebidae)', *Ethology*, vol. 106, no. 12, pp. 1067–1082, Dec. 2000, doi: 10.1046/j.1439-0310.2000.00629.x.

[46] S. Boinski, 'Affiliation Patterns Among Male Costa Rican Squirrel Monkeys', *Behaviour*, vol. 130, no. 3–4, pp. 191–209, 1994, doi: 10.1163/156853994X00523.

[47] J. Aschwanden, L. Gygax, B. Wechsler, and N. M. Keil, 'Social distances of goats at the feeding rack: Influence of the quality of social bonds, rank differences, grouping age and presence of horns', *Appl. Anim. Behav. Sci.*, vol. 114, no. 1–2, pp. 116–131, Nov. 2008, doi: 10.1016/j.applanim.2008.02.002.

[48] E. Z. Cameron, T. H. Setsaas, and W. L. Linklater, 'Social bonds between unrelated females increase reproductive success in feral horses', *Proc. Natl. Acad. Sci.*, vol. 106, no. 33, pp. 13850–13853, Aug. 2009, doi: 10.1073/pnas.0900639106.

[49] K. Langergraber, J. Mitani, and L. Vigilant, 'Kinship and social bonds in female chimpanzees ( *Pan troglodytes* )', *Am. J. Primatol.*, vol. 71, no. 10, pp. 840–851, Oct. 2009, doi: 10.1002/ajp.20711.
[50] X. Huang and S.-I. Izumi, 'Neural Alterations in Interpersonal Distance (IPD) Cognition and Its Correlation with IPD Behavior: A Systematic Review', *Brain Sci.*, vol. 11, no. 8, p. 1015, Jul. 2021, doi: 10.3390/brainsci11081015.
[51] T. P. Munyon, *An investigation of interpersonal distance and relationship quality at work*. in The Florida State University. Tallahassee, FL, USA: ProQuest Dissertations, 2009.
[52] K. S. Haring, Y. Matsumoto, and K. Watanabe, 'How do people perceive and trust a lifelike robot', in *Proceedings of the world congress on engineering and computer science*, 2013, pp. 425–430.
[53] M. Obaid, E. B. Sandoval, J. Zlotowski, E. Moltchanova, C. A. Basedow, and C. Bartneck, 'Stop! That is close enough. How body postures influence human-robot proximity', in *2016 25th IEEE International Symposium on Robot and Human Interactive Communication (RO-MAN)*, New York, NY, USA: IEEE, Aug. 2016, pp. 354–361. doi: 10.1109/ROMAN.2016.7745155.
[54] P. Fossati, 'Neural correlates of emotion processing: from emotional to social brain', *Eur. Neuropsychopharmacol.*, vol. 22, pp. S487–S491, 2012.
[55] J. A. Coan and D. A. Sbarra, 'Social Baseline Theory: the social regulation of risk and effort', *Curr. Opin. Psychol.*, vol. 1, pp. 87–91, Feb. 2015, doi: 10.1016/j.copsyc.2014.12.021.
[56] S. Schnall, K. D. Harber, J. K. Stefanucci, and D. R. Proffitt, 'Social Support and the Perception of Geographical Slant', *J. Exp. Soc. Psychol.*, vol. 44, no. 5, pp. 1246–1255, Sep. 2008, doi: 10.1016/j.jesp.2008.04.011.
[57] Edward T Hall, 'The Hidden Dimension', 1966, [Online]. Available: https://protect-eu.mimecast.com/s/f3thCV5X8c2qMy4UGZXGOU?domain=scholar.google.com Hidden Dimension
[58] R. Wolter, V. Stefanski, and K. Krueger, 'Parameters for the Analysis of Social Bonds in Horses', *Animals*, vol. 8, no. 11, p. 191, Oct. 2018, doi: 10.3390/ani8110191.
[59] D. Uzzell and N. Horne, 'The influence of biological sex, sexuality and gender role on interpersonal distance', *Br. J. Soc. Psychol.*, vol. 45, no. 3, pp. 579–597, Sep. 2006, doi: 10.1348/014466605X58384.
[60] R. Mead, A. Atrash, and M. J. Matarić, 'Automated Proxemic Feature Extraction and Behavior Recognition: Applications in Human-Robot Interaction', *Int. J. Soc. Robot.*, vol. 5, no. 3, pp. 367–378, Aug. 2013, doi: 10.1007/s12369-013-0189-8.
[61] M. L. Walters, M. A. Oskoei, D. S. Syrdal, and K. Dautenhahn, 'A long-term Human-Robot Proxemic study', in *2011 RO-MAN*, Atlanta, GA, USA: IEEE, Jul. 2011, pp. 137–142. doi: 10.1109/ROMAN.2011.6005274.
[62] T. van Oosterhout and A. Visser, 'A visual method for robot proxemics measurements', in *Proceedings of Metrics for Human-Robot Interaction: A Workshop at the Third ACM/IEEE International Conference on Human-Robot Interaction (HRI 2008). Citeseer*, Citeseer, 2008, pp. 61–68.
[63] R. Mead and M. Mataric, 'Robots Have Needs Too: How and Why People Adapt Their Proxemic Behavior to Improve Robot Social Signal Understanding', *J. Hum.-Robot Interact.*, vol. 5, no. 2, p. 48, Sep. 2016, doi: 10.5898/JHRI.5.2.Mead.
[64] B. Leichtmann and V. Nitsch, 'How much distance do humans keep toward robots? Literature review, meta-analysis, and theoretical considerations on personal space in human-robot interaction', *J. Environ. Psychol.*, vol. 68, p. 101386, Apr. 2020, doi: 10.1016/j.jenvp.2019.101386.
[65] R. Dale, G. A. Bryant, J. H. Manson, and M. M. Gervais, 'Body synchrony in triadic interaction', *R. Soc. Open Sci.*, vol. 7, no. 9, p. 200095, Sep. 2020, doi: 10.1098/rsos.200095.
[66] R. Feldman, 'Parent?infant synchrony and the construction of shared timing; physiological precursors, developmental outcomes, and risk conditions', *J. Child Psychol. Psychiatry*, vol. 48, no. 3–4, pp. 329–354, Mar. 2007, doi: 10.1111/j.1469-7610.2006.01701.x.
[67] G. R. Semin and E. R. Smith, *Embodied grounding: Social, cognitive, affective, and neuroscientific approaches*. Cambridge University Press, 2008.
[68] R. Feldman, 'Synchrony and the neurobiological basis of social affiliation.', in *Mechanisms of social connection: From brain to group.*, M. Mikulincer and P. R. Shaver, Eds., Washington: American Psychological Association, 2014, pp. 145–166. doi: 10.1037/14250-009.
[69] P. Reddish, R. Fischer, and J. Bulbulia, 'Let's dance together: Synchrony, shared intentionality and cooperation', *PloS One*, vol. 8, no. 8, p. e71182, 2013.
[70] S. Koehne, A. Hatri, J. T. Cacioppo, and I. Dziobek, 'Perceived interpersonal synchrony increases empathy: Insights from autism spectrum disorder', *Cognition*, vol. 146, pp. 8–15, Jan. 2016, doi: 10.1016/j.cognition.2015.09.007.

[71] S. S. Wiltermuth and C. Heath, 'Synchrony and cooperation', *Psychol. Sci.*, vol. 20, no. 1, pp. 1–5, 2009.

[72] M. Riehle, J. Kempkensteffen, and T. M. Lincoln, 'Quantifying facial expression synchrony in face-to-face dyadic interactions: Temporal dynamics of simultaneously recorded facial EMG signals', *J. Nonverbal Behav.*, vol. 41, pp. 85–102, 2017.

[73] K. Fujiwara, M. Kimura, and I. Daibo, 'Rhythmic Features of Movement Synchrony for Bonding Individuals in Dyadic Interaction', *J. Nonverbal Behav.*, vol. 44, no. 1, pp. 173–193, Mar. 2020, doi: 10.1007/s10919-019-00315-0.

[74] L. Gebauer, M. A. G. Witek, N. C. Hansen, J. Thomas, I. Konvalinka, and P. Vuust, 'Oxytocin improves synchronisation in leader-follower interaction', *Sci. Rep.*, vol. 6, no. 1, p. 38416, Dec. 2016, doi: 10.1038/srep38416.

[75] J. Launay, B. Tarr, and R. I. M. Dunbar, 'Synchrony as an Adaptive Mechanism for Large-Scale Human Social Bonding', *Ethology*, vol. 122, no. 10, pp. 779–789, Oct. 2016, doi: 10.1111/eth.12528.

[76] E. W. Lindsey and Y. M. Caldera, 'Shared Affect and Dyadic Synchrony Among Secure and Insecure Parent-Toddler Dyads: Shared Affect and Dyadic Synchrony', *Infant Child Dev.*, vol. 24, no. 4, pp. 394–413, Jul. 2015, doi: 10.1002/icd.1893.

[77] L. K. Miles, L. K. Nind, Z. Henderson, and C. N. Macrae, 'Moving memories: Behavioral synchrony and memory for self and others', *J. Exp. Soc. Psychol.*, vol. 46, no. 2, pp. 457–460, Mar. 2010, doi: 10.1016/j.jesp.2009.12.006.

[78] R. E. Griffioen, S. Steen, T. Verheggen, M. Enders‑Slegers, and R. Cox, 'Changes in behavioural synchrony during dog‑assisted therapy for children with autism spectrum disorder and children with Down syndrome', *J. Appl. Res. Intellect. Disabil.*, vol. 33, no. 3, pp. 398–408, May 2020, doi: 10.1111/jar.12682.

[79] F. Pirrone, A. Ripamonti, E. C. Garoni, S. Stradiotti, and M. Albertini, 'Measuring social synchrony and stress in the handler-dog dyad during animal-assisted activities: A pilot study', *J. Vet. Behav.*, vol. 21, pp. 45–52, Sep. 2017, doi: 10.1016/j.jveb.2017.07.004.

[80] 'Furhat Robotics'. https://furhatrobotics.com/ (accessed Aug. 21, 2023).

[81] 'Nao: The Educational Robot | United Robotics Group'. https://unitedrobotics.group/en/robots/nao (accessed Aug. 21, 2023).

[82] G. Calabrò, A. Bizzego, S. Cainelli, C. Furlanello, and P. Venuti, 'M-MS: A Multi-Modal Synchrony Dataset to Explore Dyadic Interaction in ASD', in *Progresses in Artificial Intelligence and Neural Systems*, A. Esposito, M. Faundez-Zanuy, F. C. Morabito, and E. Pasero, Eds., in Smart Innovation, Systems and Technologies, vol. 184. Singapore: Springer Singapore, 2021, pp. 543–553. doi: 10.1007/978-981-15-5093-5_46.

[83] M. J. Henry, P. F. Cook, K. De Reus, V. Nityananda, A. A. Rouse, and S. A. Kotz, 'An ecological approach to measuring synchronization abilities across the animal kingdom', *Philos. Trans. R. Soc. B Biol. Sci.*, vol. 376, no. 1835, p. 20200336, Oct. 2021, doi: 10.1098/rstb.2020.0336.

[84] J. Engel and J. Lamprecht, 'Doing What Everybody Does? A Procedure for Investigating Behavioural Synchronization', *J. Theor. Biol.*, vol. 185, no. 2, pp. 255–262, Mar. 1997, doi: 10.1006/jtbi.1996.0359.

[85] Y. Hu, X. Cheng, Y. Pan, and Y. Hu, 'The intrapersonal and interpersonal consequences of interpersonal synchrony', *Acta Psychol. (Amst.)*, vol. 224, p. 103513, Apr. 2022, doi: 10.1016/j.actpsy.2022.103513.